\documentclass{article}
\date{}
\usepackage{graphicx}
\usepackage[dvipsnames]{xcolor}
\usepackage{amsmath}
\PassOptionsToPackage{numbers, compress}{natbib}
\usepackage[sort&compress]{natbib}
\newcommand{\cind}[0]{\mathrel{\text{\scalebox{1.07}{$\perp\mkern-10mu\perp$}}}}
\newcommand{\ncind}[0]{\mathrel{\text{\scalebox{1.07}{$\not\perp\mkern-10mu\perp$}}}}
\usepackage{tikz, pgfplots}

\definecolor{rosso}{RGB}{220,57,18}
\definecolor{giallo}{RGB}{255,153,0}
\definecolor{blu}{RGB}{102,140,217}
\definecolor{verde}{RGB}{16,150,24}
\definecolor{viola}{RGB}{153,0,153}

\makeatletter

\tikzstyle{chart}=[
    legend label/.style={font={\scriptsize},anchor=west,align=left},
    legend box/.style={rectangle, draw, minimum size=5pt},
    axis/.style={black,semithick,->},
    axis label/.style={anchor=east,font={\tiny}},
]

\tikzstyle{bar chart}=[
    chart,
    bar width/.code={
        \pgfmathparse{##1/2}
        \global\let\bar@w\pgfmathresult
    },
    bar/.style={very thick, draw=white},
    bar label/.style={font={\bf\small},anchor=north},
    bar value/.style={font={\footnotesize}},
    bar width=.75,
]

\tikzstyle{pie chart}=[
    chart,
    slice/.style={line cap=round, line join=round, very thick,draw=white},
    pie title/.style={font={\bf}},
    slice type/.style 2 args={
        ##1/.style={fill=##2},
        values of ##1/.style={}
    }
]

\pgfdeclarelayer{background}
\pgfdeclarelayer{foreground}
\pgfsetlayers{background,main,foreground}

\newcommand{\pie}[3][]{
    \begin{scope}[#1]
    \pgfmathsetmacro{\curA}{90}
    \pgfmathsetmacro{\r}{1}
    \def\c{(0,0)}
    \node[pie title] at (90:1.3) {#2};
    \foreach \v/\s in{#3}{
        \pgfmathsetmacro{\deltaA}{\v/100*360}
        \pgfmathsetmacro{\nextA}{\curA + \deltaA}
        \pgfmathsetmacro{\midA}{(\curA+\nextA)/2}

        \path[slice,\s] \c
            -- +(\curA:\r)
            arc (\curA:\nextA:\r)
            -- cycle;
        \pgfmathsetmacro{\d}{max((\deltaA * -(.5/50) + 1) , .5)}

        \begin{pgfonlayer}{foreground}
        \path \c -- node[pos=\d,pie values,values of \s]{$\v\%$} +(\midA:\r);
        \end{pgfonlayer}

        \global\let\curA\nextA
    }
    \end{scope}
}

\begin{document}

\title{Causal Reasoning for Algorithmic Fairness}

\author{
Joshua R. Loftus$^{1}$, Chris Russell$^{2,5}$, Matt J. Kusner$^{3,5}$, and Ricardo Silva$^{4,5}$\\
$^{1}$New York University
  $^{2}$University of Surrey
    $^{3}$University of Warwick\\ 
  $^{4}$University College London
  $^{5}$Alan Turing Institute 
}

  




\maketitle

\begin{abstract}
In this work, we argue for the importance of causal reasoning in creating fair
algorithms for decision making. We give a review of existing
approaches to fairness, describe work in causality necessary for the
understanding of causal approaches, argue why causality is necessary for any approach that wishes to be fair, and give a detailed analysis of the many recent approaches to causality-based fairness.   
\end{abstract}

\section{Introduction}
The success of machine learning algorithms has created a wave of excitement
about the problems they could be used to solve. Already we have algorithms that
match or outperform humans in non-trivial tasks such as image classification \citep{he2015delving}, the game of Go \citep{silver2016mastering}, and skin cancer classification \citep{esteva2017dermatologist}. This has spurred the use of machine learning algorithms in predictive policing \citep{lum2016predict}, in loan lending
\citep{hardt2016equality}, and to predict whether released people from jail will
re-offend \citep{chouldechova2017fair}. In these life-changing settings however,
it has quickly become clear that machine learning algorithms can unwittingly
perpetuate or create discriminatory decisions that are biased against certain individuals (for example, against a particular race, gender, sexual
orientation, or other protected attributes). Specifically, such biases have already been
demonstrated in natural language processing systems
\citep{bolukbasi2016man} (where algorithms associate men with technical
occupations like `computer programmer' and women with domestic occupations like
`homemaker'), and in online advertising \citep{sweeney2013discrimination} (where
Google showed advertisements suggesting that a person had been arrested when
that person had a name more often associated with black individuals).

As machine learning is deployed in an increasingly wide range of human 
scenarios, it is more important than ever to understand what biases are present
in a decision making system, and what constraints we can put in place to
guarantee that a learnt system never exhibits such biases. Research into these
problems is referred to as \emph{algorithmic fairness}. It is a particularly
challenging area of research for two reasons: many different features are
intrinsically linked to protected classes such as race or gender. For example, in many
scenarios, knowledge of someone's address makes it easy to predict their race
with relatively high accuracy; while their choice of vocabulary might reveal much
about their upbringing or gender. As such it is to easy to accidentally create algorithms that
make decisions without knowledge of a persons race or gender, but still
exhibit a racial or gender bias.  The second issue is more challenging still,
there is fundamental disagreement in the field as to what \emph{algorithmic
  fairness} really means. 
Should algorithms be fair if they always make similar decisions for similar individuals?
Should we
instead call algorithms that make beneficial decisions for all genders at roughly the same
rate fair? Or should we use a third different criteria? This question is of
fundamental importance as many of these different criteria can not be satisfied
at the same time \cite{kleinberg2016inherent}.

In this work we argue that it is important to understand where these
sources of bias come from in order to rectify them, and that causal reasoning is
a powerful tool for doing this. We review existing notions of fairness in prediction problems; the
tools of causal reasoning; and show how these can be combined together
using techniques such as counterfactual fairness \cite{kusner:17}.

\section{Current Work in Algorithmic Fairness}
\label{sec:fairness}
To discuss the existing measures of fairness, we use capital letters to refer to
variables and lower case letters to refer to a value a variable takes. For example,
we will always use $A$ for a protected attribute such as gender, and $a$ or $a'$
to refer to the different values the attribute can take such as {\em
  man} or {\em woman}. We use $Y$ to refer to the true state of a variable we
wish to predict, for example the variable might denote whether a person defaults
on a loan or if they will violate parole conditions. We will use $\hat Y$ to denote our
prediction of the true variable $Y$. The majority of definitions of fairness in prediction problems are statements about probability of a particular prediction occurring given that some prior conditions hold. In what follows, we will use
$P(\cdot\ |\ \cdot)$ to represent either conditional probability of events, probability mass functions or density functions, as required by the context.

\subsubsection{Equalised Odds}
Two definitions of fairness that have received much attention are equalised odds
and calibration. Both were heavily used in the ProPublica investigation into Northpointe's COMPAS score, designed
to gauge the propensity of a prisoner to re-offend upon release
\citep{flores2016false}. The first measure is equalised odds, which says that if
a person truly has state $y$, the classifier will predict this at the same rate regardless
of the value of their protected attribute.  This can be written as an equation in
the following form:
\begin{align}
P(\hat{Y} = y\ |\ A = a, Y = y) = P(\hat{Y} = y\ |\ A = a', Y = y) \label{eq:eo}
\end{align}
for all $y,a,a'$. Another way of stating this property is by saying that $\hat Y$ is independent of $A$ given $Y$, which we will denote by $\hat Y \cind A\ |\ Y$.  
\subsubsection{Calibration}
The second condition is referred to as calibration (or `test fairness' in
\cite{chouldechova2017fair}). This reverses the previous condition of equalised
odds, and says that if the classifier predicts that a person has state $y$,
their probability of actually having state $y$ should be the same for all
choices of attribute.
\begin{align}
P(Y = y\ |\ A = a, \hat{Y} = y) = P(Y = y\ |\ A = a', \hat{Y} = y) \label{eq:ca}
\end{align}
for all choices of $y,a$, and $a'$, that is, $Y \cind A \ |\ \hat{Y}$.

Although the two measures sound very similar, they are fundamentally
incompatible. These two measures achieved some notoriety when Propublica showed
that Northpointe's COMPAS score violated equalised odds, accusing them of
racial discrimination. In response, Northpointe claimed that their COMPAS
score satisfied calibration and that they did not discriminate. Kleinberg et al.
\citep{kleinberg2016inherent} and Chouldechova \citep{chouldechova:17}
showed that both conditions cannot be
satisfied at the same time except in special cases such as zero prediction error
or if $Y \cind A$.
     
The use of calibration and equalised odds has another major
limitation. If $Y \ncind A$, the true scores $Y$ typically have some inherent bias. This happens, for example, if the 
police are more likely to unfairly decide that minorities are violating
their parole. The definitions of calibration or equalised odds do not explicitly forbid the classifier from preserving an existing bias. 

\subsubsection{Demographic Parity/Disparate Impact}
Perhaps the most common  non-causal notion of fairness is \emph{demographic parity}, defined as follows:
\begin{align}
P(\hat{Y} = y\ |\ A = a) = P(\hat{Y} = y\ |\ A = a'), \label{eq:dp}
\end{align}
for all $y,a,a'$, that is, $\hat Y \cind A$. If unsatisfied, this notion is also referred to as \emph{disparate impact}. Demographic parity has been used, for several purposes, in the following works:
\cite{kamiran2009classifying, kamishima2012fairness, zemel2013learning,
  louizos2015variational, edwards2015censoring, zafar2017fairness}.

Satisfying demographic parity can often
require positive discrimination, where certain individuals who
are otherwise very similar are treated differently due to having different
protected attributes. Such \emph{disparate treatment} can violate other intuitive notions of fairness or equality, contradict equalised odds or calibration, and in some cases is prohibited by law.

\subsubsection{Individual Fairness}
Dwork et al. \cite{dwork2012fairness} proposed the concept of \emph{individual fairness} as follows.
\begin{align}
P(\hat Y^{(i)} = y\ |\ X^{(i)}, A^{(i)}) \approx
   P(\hat Y^{(j)} = y\ |\ X^{(j)}, A^{(j)}), \mbox{ if } d(i, j) \approx 0, \label{eq:if}
\end{align}
where $i,j$ refer to two different individuals and the superscripts $(i),(j)$
are their associated data. The function $d(\cdot, \cdot)$ is a `task-specific'
metric that describes how any pair of individuals should be treated similarly in
a fair world. The work suggests that this metric could be defined by `a
regulatory body, or \ldots a civil rights organization'. While this notion
mitigates the issues with individual predictions that arose from demographic
parity, it replaces the problem of defining fairness with defining a fair metric
$d(\cdot, \cdot)$. As we observed in the introduction, many variables vary along with
protected attributes such as race or gender, making it challenging to
find a distance measure that will not allow some implicit discrimination.

\subsubsection{Causal Notions of Fairness}
A number of recent works use causal approaches to address fairness 
\cite{kusner:17, elias:18, silvia:18, pmlr-v81-barabas18a, russell:17, kilbertus:17}, which we review in more detail
in Section \ref{sec:caus-noti-fairn}. We describe selected background on
causal reasoning in Section \ref{sec:causal}. These works depart from the previous
approaches in that they are not wholly data-driven but require additional
knowledge of the structure of the world, in the form of a causal model.  
This additional knowledge is particularly valuable as it informs us how
changes in variables propagate in a system, be it natural, engineered or social. Explicit causal assumptions remove ambiguity from methods that just depend upon
statistical correlations. For instance, causal methods provide a recipe to express assumptions on how to recover from sampling biases in the data (Section \ref{sec:why_causality}) or how to describe mixed scenarios where we may believe that certain forms of discrimination should be allowed while others should not (e.g., how gender influences one's field of study in college, as in Section \ref{sec:caus-noti-fairn}). 

\section{Causal Models}
\label{sec:causal}
We now review causality in sufficient detail for our analysis of causal fairness
in Section \ref{sec:caus-noti-fairn}.  
It is challenging to give a self-contained definition of causality, as many working definitions reveal 
circularities on close inspection. For two random variables $X$ and $Y$, informally we say
that $X$ \emph{causes} $Y$ when there exist at least two different
\emph{interventions} on $X$ that result in two different probability distributions
of $Y$. This does not mean we will be able to define what an
``intervention'' is without using causal concepts, hence circularities
appear.

Nevertheless, it is possible to formally express causal assumptions
and to compute the consequences of such assumptions if one is willing
to treat some concepts, such as interventions, as primitives. This is just an instance of the
traditional axiomatic framework of mathematical modelling, dating back to
Euclid. In particular, in this paper we will make use primarily of the
\emph{structural causal model} (SCM) framework advocated by
\cite{pearl:00}, which shares much in common with the approaches by
 \cite{robins:86} and \cite{sgs:93}. 

\subsection{Structural Causal Models}

We define a causal model as a triplet $(U, V, F)$ of sets such that:
\begin{itemize}
\item $V$ is a set of observed random variables that form the causal system of our interest;
\item $U$ is a set of latent (that is, unobservable) {\bf background} variables that will
  represent all possible causes of $V$ and which jointly follow some distribution $P(U)$;
\item $F$ is a set of functions $\{f_1, \dots, f_n\}$, one for each
  $V_i \in V$, such that $V_i = f_i(pa_i, U_{pa_i})$, $pa_i \subseteq
  V \backslash \{V_i\}$ and $U_{pa_i} \subseteq U$. Such equations are
  also known as {\bf structural equations} \cite{bol:89}.
\end{itemize}
The notation $pa_i$ is meant to capture the notion that a directed
graph $\mathcal G$ can be used to represent the input-output
 relationship encoded in the structural equations: each
vertex $X$ in $\mathcal G$ corresponds to one random variable
in $V \cup U$, with the same symbol used to represent both the vertex
and the random variable; an edge $X \rightarrow V_i$ is added to
$\mathcal G$ if $X$ is one of the arguments of $V_i \!=\! f_i(\cdot)$. Hence,
$X$ is said to be a \emph{parent} of $V_i$ in $\mathcal G$.  In what
follows, we will assume without loss of generality that vertices in
$U$ have no parents in $\mathcal G$. We will also assume that
$\mathcal G$ is acyclic, meaning that it is not possible to start from
a vertex $X$ and return to it following the direction of the edges.

A SCM is causal in the sense it allows us to predict \emph{effects of
  causes} and to infer \emph{counterfactuals}, as discussed below.

\subsection{Interventions, Counterfactuals and Predictions}

The effect of a cause follows from an operational notion of
intervention. This notion says that a \emph{perfect intervention} on a
variable $V_i$, at value $v$, corresponds to overriding $f_i(\cdot)$
with the equation $V_i = v$. Once this is done, the joint
distribution of the remaining variables $V_{\backslash i} \equiv V
\backslash \{V_i\}$ is given by the causal model. Following \cite{pearl:00}, we will
denote this operation as $P(V_{\backslash i}\ |\ do(V_i = v_i))$. This notion is
immediately extendable to a set of simultaneous interventions on multiple
variables.

The introduction of the $do(\cdot)$ operator emphasises the difference between
$P(V_{\backslash i}\ |\ do(V_i = v_i))$ and $P(V_{\backslash i}\ |\ V_i = v_i)$. As
an example, consider the following structural equations:
\begin{eqnarray}
    Z &=& U_Z, \label{eq:main1}\\
    A &=& \lambda_{az}Z + U_A, \label{eq:main2}\\
    Y &=& \lambda_{ya}A + \lambda_{yz}Z + U_Y. \label{eq:main3}
\end{eqnarray}

\begin{figure}[!t]
\centering\includegraphics[width=\textwidth]{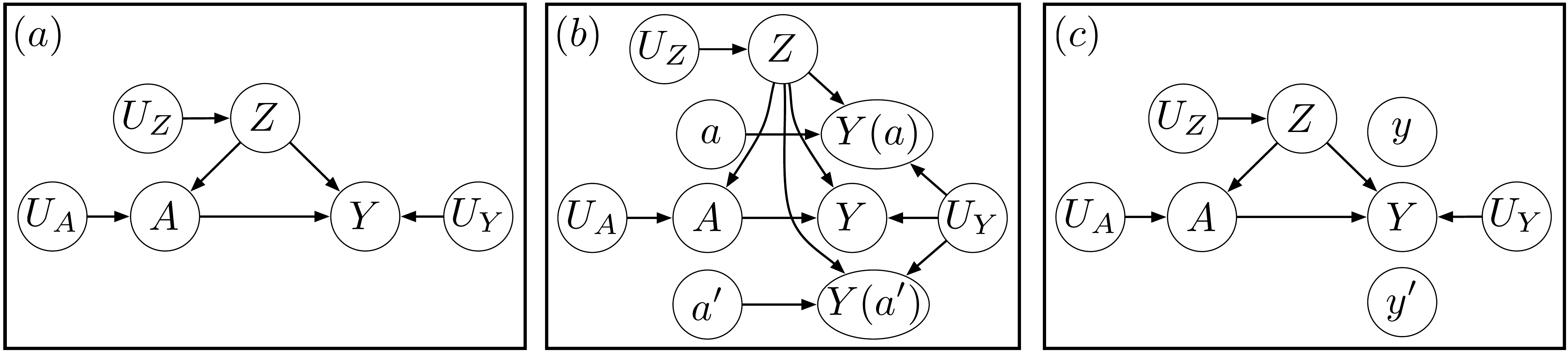}
  \caption{(a) A causal graph for three observed variables $A, Y, Z$.
    (b) A joint representation with explicit background variables, and
    two counterfactual alternatives where $A$ is intervened at two
    different levels.  (c) Similar to (b), where the interventions
    take place on $Y$.}
\label{fig:basic_figure}
\end{figure}

The corresponding graph is shown in Figure \ref{fig:basic_figure}(a).
Assuming that the background variables follow a standard Gaussian with
diagonal covariance matrix, standard algebraic manipulations
allows us to calculate that $P(Y = y\ |\ A = a)$ has a Gaussian density with a mean that depends on
$\lambda_{az}, \lambda_{ya}$ and $\lambda_{yz}$. In contrast, $\mathsf
E[Y\ |\ do(A = a)] = \lambda_{ya}a$, which can be obtained by first
erasing (\ref{eq:main2}) and replacing $A$ with $a$ on the
right-hand side of (\ref{eq:main3}) followed by marginalizing the remaining variables. The difference illustrates the dictum ``causation is not
correlation'': $Z$ acts as a \emph{confounder} (common cause) of
\emph{exposure} $A$ and \emph{outcome} $Y$. In a randomised controlled
trial (RCT), $A$ is set by design, which breaks its link with $Z$.
In an observational study, data is generated by the system above, and
standard measures of correlation between $A$ and $Y$ will not provide
the correct interventional distribution: $P(Y\ |\ do(A = a))$.
The $do(\cdot)$ operator captures the notion of \emph{effect of a
  cause}, typically reported in terms of a contrast such as $\mathsf
E[Y\ |\ do(A = a)] - \mathsf E[Y\ |\ do(A = a')]$ for two different
intervention levels $a, a'$.

Another causal inference task is the computation of
\emph{counterfactuals} implied from causal assumptions and
observations: informally, these are outcomes following from
alternative interventions on the same \emph{unit}. A ``unit'' is the
snapshot of a system at a specific context, such as a person at a
particular instant in time. Operationally, a unit can be understood as
a particular instantiation of the background variable set $U$, which
determine all variables in $V$ except for those being intervened
upon. Lower-case $u$ will be used to represent such realisations,
with $U$ interpreted as a random unit. The name
``counterfactual'' comes from the understanding that, if the
corresponding exposure already took place, then any such alternative
outcomes would be (in general) contrary to the realised facts. Another
commonly used term is \emph{potential outcomes} \cite{rubin:74}, a
terminology reflecting that strictly speaking such outcomes are not
truly counterfactual until an exposure effectively takes place.

For any possible intervention $V_j = v_j$ for unit $u$, we denote the
counterfactual of $V_i$ under this intervention as $V_i(v_j, u)$. This
notation also accommodates the simultaneous hypothetical interventions
on the corresponding set of background variables $U$ at level $u$.
The factual realisation of a random variable $V_i$ for unit $u$ is
still denoted by $V_i(u)$. The random counterfactual corresponding to
intervention $V_j = v_j$ for an unspecified unit $U$ is denoted as
$V_i(v_j, U)$ or, equivalently, $V_i(v_j)$ for notational simplicity. 

By treating $U$ as a set of random variables, this implies that
factuals and counterfactuals have a joint distribution. One way of
understanding it is via Figure \ref{fig:basic_figure}(b), which
represents a factual world and two parallel worlds where $A$ is set to
intervention levels $a$ and $a'$. A joint distribution for $Y(a)$ and $Y(a')$
is implied by the model. Conditional distributions, such as $P(Y(a) = y_a,
Y(a') = y_{a'}\ |\ A = a, Y = y, Z = z)$ are also defined. Figure \ref{fig:basic_figure}(c)
shows the case for interventions on $Y$. It is not difficult to show,
as $Y$ is not an ancestor of $A$ in the graph, that $A(y, u) = A(y', u) =
A(u)$ for all $u, y, y'$. This captures the notion the $Y$ does not cause $A$.

\subsection{Counterfactuals Require Untestable Assumptions}
\label{sec:untestable}
Unless structural equations depend on observed variables only, they
cannot be tested for correctness (unless other untestable assumptions
are imposed). We can illustrate this problem by noting that a
conditional density function $P(V_j\ |\ V_i = v)$ can be written as an
equation $V_j = f_1(v, U) \equiv F^{-1}_{V_i = v}(U) = F^{-1}_{V_i = v}(g^{-1}(g(U))) \equiv f_2(v, U')$,
where $F^{-1}_{V_i = V}(\cdot)$ is the inverse cumulative distribution
function corresponding to $P(V_j\ |\ V_i = v)$, $U$ is an uniformly distributed
random variable on $[0, 1]$, $g(\cdot)$ is some
arbitrary invertible function on $[0, 1]$, and $U' \equiv g(U)$. While this is not
fundamental for effects of causes, which depend solely on predictive
distributions 
that at least in
theory can be estimated from RCTs, different structural equations with
the same interventional distributions will imply different joint
distributions over the counterfactuals.

The traditional approach for causal inference in statistics tries to
avoid any estimand that cannot be expressed by the marginal
distributions of the counterfactuals (i.e., all estimands in which
marginals $P(Y(a) = y_a)$ and $P(Y(a') = y_{a'})$ would provide enough information, such as the
\emph{average causal effect} $\mathsf{E}[Y(a) - Y(a')] =
\mathsf{E}[Y\ |\ do(A = a)] - \mathsf{E}[Y\ |\ do(A = a')]$). Models that follow this approach and specify solely the univariate marginals of a counterfactual joint
distribution are sometimes called \emph{single-world} models
\cite{richardson:13}. However, as we will see, \emph{cross-world}
models seem a natural fit to algorithmic fairness. In particular, they
are required for non-trivial statements that concern fairness at an individual
level as opposed to fairness measures averaged over groups of
individuals.

\section{Why Causality is Critical For Fairness}
\label{sec:why_causality}
Ethicists and social choice theorists recognise the importance of causality
in defining and reasoning about fairness. Terminology varies,
but many of their central questions and ideas, such as the role of agency
in justice, responsibility-sensitive egalitarianism, and luck egalitarianism
\cite{richard1988equality, cohen1989currency, dworkin2002sovereign}
involve causal reasoning. Intuitively, it is unfair
for individuals to experience different outcomes caused by factors
outside of their control.
Empirical studies
of attitudes about distributive justice
\cite{cappelen2013just, mollerstrom2015luck}
have found that most participants prefer redistribution to create
fairer outcomes, and do so in ways that depend on how much control
individuals have on their outcomes.
Hence, when choosing policies and designing systems
that will impact people, we should minimise or eliminate the
causal dependence on factors outside an individual's control, such as
their perceived race
or where they were born. Since such factors have influences on other
aspects of peoples' lives that may also be considered relevant for
determining what is fair, applying this intuitive notion of fairness
requires careful causal modelling as we describe here.

Is it necessary that models attempting to remove such factors be causal? Many
other notions of algorithmic fairness have also attempted to control or adjust
for covariates. While it is possible to produce identical predictions or
decisions with a model that is equivalent mathematically but without overt
causal assumptions or interpretations, the design decisions underlying
a covariate adjustment are often based on implicit causal reasoning. There is a fundamental benefit from an explicit statement of these assumptions. To illustrate this, we consider a classic example of bias in graduate admissions.

\subsection{Revisiting Gender Bias In Berkeley Admissions}

The Berkeley admissions example \cite{bickel1975sex} is often used to
explain Simpson's paradox 
\cite{simpson1951interpretation} and highlight the importance of adjusting
for covariates. In the fall of 1973, about 34.6\% of women and 44.3\% of
men who applied to graduate studies at Berkeley were admitted. However,
this was not evidence that the admissions decisions were biased against
women. Decisions were made on a departmental basis, and each department
admitted proportions of men and women at approximately the same rate.
However, a greater proportion of women applied to the most selective departments,
resulting in a lower overall acceptance rate for women.

While the overall outcome is seemingly unfair, after controlling for choice of
department it appears to be fair, at least in some sense. In fact,
while the presentation of this example to illustrate Simpson's paradox
often ends there, the authors in \cite{bickel1975sex} conclude,
"Women are shunted by their socialisation and education toward fields 
of graduate study that are generally more crowded, less productive of 
completed degrees, and less well funded, and that frequently offer
poorer professional employment prospects." The outcome can still be judged to be unfair, not due to biased admissions decisions, but rather to the causes of
differences in choice of department, such as socialisation. Achieving 
or defining fairness requires addressing those root causes and applying
value judgements.  Applicants certainly have some agency over which department they apply to, but that decision is not made free of outside influences. They
had no control over what kind of society they had been born into,
what sort of gender norms that society had during their lifetime,
or the scarcity of professional role models, and so on. 

The quote above suggests that the authors in \cite{bickel1975sex} were
reasoning about causes even if they did not make  explicit use of causal
modelling. Indeed, conditioning on the choice of the department only makes
sense because we understand it has a causal relationship with the outcome
of interest and is not just a spurious correlation. Pearl \cite{pearl:00} provides a detailed account of the causal basis of Simpson's paradox.

\subsection{Selection Bias and Causality}
\label{sec:select-bias-caus}
Unfairness can also arise from bias in how data is collected or
sampled. For instance, if the police stop individuals on the street to
check for the possession of illegal drugs, and if the stopping protocol is the
result of discrimination that targets individuals of a particular race, this can
create a feedback loop that justifies discriminatory practice. Namely, if
data gathered by the police suggests that $P(Drugs = yes\ |\ Race = a) > P(Drugs = yes\ |\ Race
= a')$, this can be exploited to justify an unbalanced stopping process when
police resources are limited. How then can we assess its fairness? It is possible to
postulate structures analogous to the Berkeley example, where a mechanism such
as $Race \rightarrow Economic\ status \rightarrow Drugs$ 
explains the pathway. Debate would focus on the level of agency of an individual
on finding himself or herself at an economic level that leads to increased
drug consumption.

But selection bias cuts deeper than that, and more recently causal
knowledge has been formally brought in to understand the role of such biases \cite{spirtes:95b,elias:16}. This is achieved by representing a selection variable $Selected$ as part of our model, 
and carrying out inference by acknowledging that, in the data, all individuals are such that ``$Selected = true$''. The association between race and drug is expressed as
$P(Drugs = yes\ |\ Race = a, Selected = true) > P(Drugs = yes\ |\ Race = a', Selected = true)$,
which may or may not be representative of the hypothetical population in which everyone has been examined. The data cannot directly tell whether
$P(Drugs = yes\ |\ Race = a, do(Selected = true)) > P(Drugs = yes\ |\ Race = a', do(Selected = true))$.
As an example, it is possible to postulate the two following causal structures that cannot be distinguished
on the basis of data already contaminated with selection bias: (i) the structure $Race \rightarrow Drugs$,
with $Selected$ being a disconnected vertex; (ii) the structure $Race \rightarrow Selected \leftarrow H
\rightarrow Drugs$, where $H$ represents hidden variables not formally logged in police records.

In the latter case, we can check that drugs and race and unrelated. However,
$P(Drugs = yes\ |\ Race = a, Selected = true) \neq P(Drugs = yes\ |\ Race = a',
Selected = true)$, as conditioning on $Selected$ means that both of its causes
$Race$ and $H$ ``compete'' to explain the selection. This induces an association
between $Race$ and $H$, which carries over the association between $Race$ and
$Drugs$. At the same time, $P(Drugs = yes\ |\ Race = a, do(Selected = true)) =
P(Drugs = yes\ |\ Race = a', do(Selected = true))$, a conclusion that cannot be
reached without knowledge of the causal graph or a controlled experiment making
use of interventions. Moreover, if the actual structure is a combination of (i)
and (ii), standard statistical adjustments that remove the association between
$Race$ and $Drugs$ cannot disentangle effects due to selection bias from those
due to the causal link $Race \rightarrow Drugs$, harming any arguments that can
be constructed around the agency behind the direct link.

\subsection{Fairness Requires Intervention}

Approaches to algorithmic fairness usually involve imposing some kind of
constraints on the algorithm (such as those formula given by
Section~\ref{sec:fairness}). We can view this as an intervention on the
predicted outcome $\hat Y$. And, as argued in \cite{pmlr-v81-barabas18a}, we can
also try to understand the causal implications for the system we are intervening
on. That is, we can use an SCM to model the causal relationships between
variables in the data, between those and the predictor $\hat Y$ that we are
intervening on, and between $\hat Y$ and other aspects of the system that will
be impacted by decisions made based on the output of the algorithm.

To say that fairness is an intervention is not a strong statement considering
that any decision can be considered to be an intervention. Collecting data,
using models and algorithms with that data to predict some outcome variable, and
making decisions based on those predictions are all intentional acts motivated
by a causal hypothesis about the consequences of that course of action. In
particular, \emph{not} imposing fairness can also be a deliberate intervention,
albeit one of inaction.

We should be clear that prediction problems do not tell the whole story. Breaking the causal links between $A$ and a prediction $\hat Y$ is a way of avoiding some unfairness in the world, but it is only one aspect of the problem. Ideally, we would like that no paths from $A$ to $Y$ existed, and the provision of fair predictions is predicated on the belief that it will be a contributing factor for the eventual change in the generation of $Y$. We are not, however, making any formal claims of modelling how predictive algorithmic fairness will lead to this ideal stage where causal paths from $A$ to $Y$ themselves disappear.

\section{Causal Notions of Fairness}
\label{sec:caus-noti-fairn}
In this section we discuss some of the emerging notions of fairness formulated in
terms of SCMs, focusing in particular on a notion introduced by us in
\cite{kusner:17}, \emph{counterfactual fairness}.  We explain how
counterfactual fairness relates to some of the more well-known notions
of statistical fairness and in which ways a causal perspective
contributes to their interpretation.  The remainder of the section
will discuss alternative causal notions of fairness and how
they relate to counterfactual fairness. 

\subsection{Counterfactual Fairness}

A predictor $\hat Y$ is said to satisfy \emph{counterfactual fairness} if
\begin{equation}
  P(\hat Y(a, U) = y\ |\ X = x, A = a) =  P(\hat Y(a', U) = y\ |\ X = x, A = a),
\label{eq:cf}
\end{equation}
\noindent for all $y, x, a, a'$ in the domain of the respective variables
\cite{kusner:17}. The randomness here is on $U$ (recall that background variables $U$ can be thought of as describing a particular individual person at some point in time). In practice, this
means we can build $\hat Y$ from any variable $Z$ in the system which is
not caused by $A$ (meaning there is no directed path from $A$ to $Z$ in the corresponding graph). This includes the background
variables $U$ which, if explicitly modelled, allows us to
use as much information from the existing observed variables as
possible\footnote{By conditioning on the
event $\{X = a, A = a\}$ we are essentially extracting information from the
individual to learn these background variables.}.

This notion captures the intuition by which, ``other things being equal''
(i.e., the background variables), our prediction would not have changed in the
parallel world where only $A$ would have changed.  Structural equations provide an operational meaning for counterfactuals
such as ``what if the race of the individual had been different''.
This can be interpreted solely as comparing two hypothetical individuals which are
identical in background variables $U$ but which differ in the way $A$ is set.
As described in Section \ref{sec:causal}\ref{sec:untestable}, in most cases we will not be able to verify that the proposed structural equations perfectly describe the unfairness in the data at hand.  
Instead, they are means to describe explicit assumptions made about the mechanisms of unfairness, and to expose these assumptions openly to criticism.

The application of counterfactual fairness requires a causal model
$\mathcal M$ to be formulated, and training data for the observed
variables $X$, $A$ and the target outcome $Y$. Unlike other approaches
discussed in the sequel, we purposely avoid making use of any
information concerning the structural equation for $Y$ in model
$\mathcal M$\footnote{If $Y$ provides information for some parameters
  of $\mathcal M$, we can of course fit $\mathcal M$ using this
  data. Our point here is what happens \emph{after} $\mathcal M$ has
  been decided.}. This is motivated by the fact that $\hat Y$ must not
make use of $Y$ at test time. At training time, the only thing that
matters for any prediction algorithm of interest is to reconstruct $Y$
directly, which can be done via the data. Note that the data could
also be Monte Carlo samples drawn from a theoretical model.

As an example, consider the simple causal model $A \rightarrow X \rightarrow
Y$, $U_X \rightarrow X$,  with the structural equation $X = f_X(A, U_X)$. $\hat Y$ must not
be a function of $A$ or $X$ except via $U_X$, or otherwise $\hat Y(a,
u_X) \neq \hat Y(a', u_X)$ will be different since in general $f_X(a,
u_X) \neq f_X(a', u_X)$. For any $\hat Y \equiv g(U_X)$, marginalising
$U_X$ via $P(U_X\ |\ X = x, A = a)$ guarantees (\ref{eq:cf}). We can use any statistical method to fit $g(\cdot)$, e.g., 
by minimising some loss function with respect to the distribution of $Y$ directly.




\subsection{Counterfactual Fairness and Common Statistical Notions}
We now relate counterfactual fairness to the non-causal, statistical notions of fairness introduced in Section~\ref{sec:fairness}.

\subsubsection{Demographic Parity}
Note that if the background variables $U$ are
uniquely determined by observed data $\{X = a, A = a\}$, and $\hat{Y}$ is a
function only of background variables $U$ and observed variables which are independent of $A$, then a counterfactually fair predictor will satisfy eq.~\eqref{eq:dp}. In this sense,
counterfactual fairness can be understood to be a counterfactual analogue of
demographic parity.


\subsubsection{Individual Fairness}
If we think of the two different individuals defined in eq.~\eqref{eq:if} as \emph{matched} by a statistical matching procedure \cite{morgan:15}, then they can be thought of as counterfactual versions of each other. Specifically, in such a procedure the counterfactual version of
individual $i$ that is used to estimate a causal effect is in reality an
observed case $j$ in a sample of controls, such that $i$ and $j$
are close in some space of pre-treatment variables (that is,
variables not descendant of $A$) \cite{morgan:15}. This ``closeness'' can be directly specified by the distance metric in eq.~\eqref{eq:if}. Interpreted this way the individual fairness condition is similar to a particular instantiation of counterfactual fairness (i.e., via matching). One notable difference is that in \cite{dwork2012fairness} the fairness condition holds for all pairs of individuals, not just for the closest pairs as in matching. Unlike matching, the predictor is encouraged to be different for individuals that are not close.

\subsubsection{Equalised Odds and Calibration}
A sufficient condition for $Y \cind A$ is that there are no causal paths
between $Y$ and $A$ (this condition is also necessary under the
assumption of \emph{faithfulness} discussed by \cite{sgs:93}). In
this case, it not hard to show graphically that a counterfactually
fair $\hat Y$ built by just using non-descendants of $A$ in the
graph will respect both equalised
odds ($\hat Y \cind A\ |\ Y$) and calibration ($Y \cind A\ |\ \hat Y$). Likewise, if there exists a $\hat Y$ such that $\hat Y = Y$
for all units (zero prediction error), this can be recovered by a
causal model that postulates all the inputs to the structural
equation of $Y$, and where such inputs can be recovered from the
observed covariates.

We can also argue that if $Y \ncind A$, then neither
equalised odds, eq.~\eqref{eq:eo}, nor calibration, eq.~\eqref{eq:ca}, may be desirable. For instance,
if $A$ is an ancestor of $Y$, then we should not try to reproduce
$Y$ as it is a variable that carries bias according our counterfactual
definition (using $\hat Y = Y$). It becomes unclear why we should
strive to achieve (e.g.) calibration when our target should be
``fair'' components that explain the variability of $Y$ (like the
non-descendants of $A$) instead of all sources of variability that
generate $Y$.

\subsection{A Framework to Categorise Causal Reasoning in Fairness}

Counterfactual fairness was originally proposed as a particular
trade-off linking strength of assumptions and practicalities. In
parallel, other notions have emerged which propose different
trade-offs. In what follows, we provide a critical appraisal of alternatives and a
common framework to understand the dimensions in which causal
reasoning plays a role in algorithmic fairness. The dimensions are:
\begin{itemize}
\item \emph{individual vs. group level causal effects.} As discussed
  in Section \ref{sec:causal}, counterfactual reasoning takes place
  at the individual level, while distributions indexed by the $do(\cdot)$
  operator are meant to capture the effects of actions in groups.
  There is a clear advantage on targeting group effects, since they
  do not require postulating an unobservable joint distribution of two
  or more outcomes which can never be observed together, where even the
  \emph{existence} of counterfactuals can justifiably be treated as
  a metaphysical concept \cite{dawid:00}. However, fairness is commonly
  understood at an individual level, where unit-level assumptions are required in many scenarios;
\item \emph{explicit vs. implicit structural equations.} Although
  counterfactual models require untestable assumptions, not all
  assumptions are created equal. In some scenarios, it is possible
  to obtain some degree of fairness by postulating independence constraints
  among counterfactuals without committing to any particular interpretation
  of latent variables. This is however not always possible without losing significant information;
\item \emph{prediction vs. explanation.} Sometimes the task is not to
  create a new fair predictor, but to quantify in which ways an existing
  decision-making process is discriminatory;
\end{itemize}

Counterfactual fairness, for instance, is a notion that (i) operates at
the individual level, (ii) has explicit structural equations and, (iii) targets
prediction problems. Besides using this framework to categorise
existing notions, we will provide a discussion on path-specific
effects and how they relate to algorithmic fairness.

\subsubsection{Purely Interventional Notions}

Due to the ultimately untestable nature of counterfactuals, it is
desirable to avoid several of its assumptions whenever possible. One
way to do this is by defining constraints on the \emph{interventional
  distributions} $P(\hat Y\ |\ do(A = a), X = x)$ only. The work by
\cite{kilbertus:17} emphasises this aspect, while making explicit
particular notions of path-specific effects which we will discuss in a
latter section. The interventional notion advanced by
\cite{kilbertus:17} is the constraint:
\begin{align*}
P(\hat Y\ |\ do(A = a)) = P(\hat Y\ |\ do(A = a')).
\end{align*}
A predictor
$\hat Y$ is constructed starting from the causal model $\mathcal M$
for the system. A family of models is constructed by
modifications to $\mathcal M$ so that the total effect of $A$ on $\hat Y$ is
cancelled. The family of models itself can be parameterised so that
minimising error with respect to $Y$ is possible within this
constrained space.

To understand this via an example, consider a variation of equations
(\ref{eq:main1})-(\ref{eq:main3}) where now we have four variables,
$Z, A, X, \hat Y$. Variable $A$ is the protected attribute, $X = \lambda_{xa}A
+ \lambda_{xz}Z + U_X$ and $\hat Y = \lambda_{\hat ya}A +
\lambda_{\hat yz}Z + \lambda_{\hat yx}X + U_Y$. Parameters $\Lambda
\equiv \{\lambda_{\hat ya}, \lambda_{\hat yz}, \lambda_{\hat yx}\}$
are free parameters, while those remaining are assumed to be part of the
given causal model. However, such free parameters will be
constrained. This can be seen by substituting the equation for $X$ into that for $\hat Y$ as follows: $\hat Y = \lambda_{\hat ya}A + \lambda_{\hat yz}Z +
\lambda_{\hat yx}(\lambda_{xa}A + \lambda_{zx}Z + U_X) + U_Y$. As
$U_X$ and $U_Y$ are assumed to be independent of $A$ by construction,
it is enough to consider the sub-models where $\lambda_{\hat ya} +
\lambda_{\hat yx}\lambda_{xa} = 0$ (i.e, the total contribution of $A$ to $\hat Y$ is $0$), optimising $\Lambda$ to
minimise the prediction error of $Y$ under this constraint.

Although it seems strange that this example and other examples in
\cite{kilbertus:17} use equations, they do not need to be interpreted
as \emph{structural} in order to verify that $P(\hat Y\ |\ do(A = a))
= P(\hat Y\ |\ do(A = a'))$. The equations are just means of
parameterising the model, with latent variables $U$ being calculation
devices rather than having a real interpretation. If we assume that
equations \emph{are structural} we would revert back to counterfactual fairness,
where the procedure above is just an indirect way of regressing on
hidden variables $U_X$. However, there are important issues with the
interventional approach as it represents group-level rather than
individual-level causal effects. This means it is perfectly possible
that $\hat Y$ is highly discriminatory in a counterfactual sense and yet satisfies the purely
interventional criterion: consider the case where the structural
equation $Y = f(A, U_Y)$ is such that $P(U_Y = 0) = P(U_Y = 1) = 1/2$
and $f(a, 1) = a$, $f(a, 0) = 1 - a$, for $a \in \{0, 1\}$. Then $P(Y
= 1\ |\ do(A = 1)) = P(Y = 1\ |\ do(A = 0)) = 1/2$, even though for
every single individual we have that $Y(a, u_Y) = 1 - Y(1 - a,
u_Y)$ (i.e., we get exactly the opposite prediction for a fixed individual $u_Y$ if their race $a$ changes). Conditioning on other attributes does not help: the expression
$P(Y\ |\ do(A = a), X = x) - P(Y\ |\ do(A = a'), X = x)) = 0$ is not realistic
if $X$ is a descendant of $A$ in the causal graph, since in this case
no single individual will keep $X$ at a fixed level as $A$
hypothetically varies. This is a comparison among different
individuals who are assigned different $A$ and yet happen to coincide on
$X$. Thus we argue that it is less obvious how to motivate this notion of
conditioning. This is not an issue if $X$ is not a descendant of $A$,
but then again, in this case, we do not need to know the structural
equation for $X$ in order to use it according to counterfactual
fairness.

\subsubsection{Counterfactuals without Explicit Structural Equations}

When the goal is to minimise the number of untestable assumptions,
one direction is to avoid making any explicit assumptions about the structural equations of the causal model.
In this line of work, assumptions about the directed edges in the causal graph and a
parametric formulation for the observed distributions are allowed,
as well as independence constraints among counterfactuals. However,
no explicit structural equations are allowed.

The work by \cite{ilya:18} presents such ideas by making a direct
connection to the long tradition in graphical causal models of
providing algorithms for identifying causal estimands. That is,
without assuming any particular parametric contribution from latent
variables, such approaches either provide a function of $P(V)$ (recall that $V$ is the set of all observed variables) that is
equal to a causal effect of interest, or report whether such a
transformation is not at all possible (that is, the causal effect of
interest is \emph{unidentifiable} from $P(V)$ and a causal graph)
\cite{pearl:00}. The main idea of \cite{ilya:18} is to first identify which causal effects of protected attribute $A$ on outcome $Y$
should be (close to) zero (i.e., those we wish to make fair). For instance, similar to counterfactual fairness,
we may require that $\mathsf E[Y(a)] = \mathsf E[Y(a')]$. In some models,
we can express $\mathsf E[Y(a)] - \mathsf E[Y(a')] = 0$ as a constraint on
$P(V)$. The model can then be fit subject to such a constraint.
Predictor $\hat Y$ is what the constrained model predicts as $Y$.

Despite its desirable features, there are major prices to be paid by
avoiding structural equations. For technical reasons, enforcing such
constraints require throwing away information from any descendant of
$A$ that is judged to be on an ``unfair path'' from $A$ to $Y$. Further, in many
cases, the causal effect of interest is not identifiable. Even if
it is, the constrained fitting procedure can be both computationally
challenging and not necessarily have a clear relationship to the actual
loss function of interest in predicting $Y$. This is because it first requires
assuming a model for $Y$ and fitting a projection of the model to the
space of constrained distributions. And while explicit claims about
structural equations are avoided, the approach still relies on
cross-world assumptions of independences among counterfactuals.

A different take with a similar motivation was recently introduced by
\cite{elias:18}.  It focuses on decomposing counterfactual measures of
causal effects across different paths to \emph{explain}
discrimination. For example, the association between $A$ and $Y$ can
be decomposed by the causal effect of $A$ on $Y$ that is due to
directed paths from $A$ to $Y$ and due to a common cause of $A$ and
$Y$. Although no particular way of building a fair predictor $\hat Y$
is provided, \cite{elias:18} explicitly discuss how modifications to
particular parts of the causal model can lead to changes in the
counterfactual measures of discrimination, an important and
complementary goal of the problem of fair predictions. This work
also illustrates alternative interpretations of causal unfairness: in
our work, we would not consider $\hat Y$ to be unfair if the causal
graph is $A \leftarrow X \rightarrow \hat Y$, as $A$ is not a cause of
$\hat Y$, while \cite{elias:18} would label it as a type of ``spurious''
discrimination.  Our take is that if $\hat Y$ is deemed unfair, it
must be so by considering $X$ as yet another protected attribute.

Regarding the challenges posed by causal modelling, we advocate that
assumptions about structural equations are still useful if interpreted
as claims about how latent variables with a clear domain-specific
meaning contribute to the pathways of interest. Even if imperfect,
this direction is a way of increasing transparency about the
motivation for labelling particular variables and paths as ``unfair''.
It still offers a chance of falsification and improvement when
then-latent variables are measured in subsequent studies. If there are
substantive disagreements about the functional form of structural
equations, multiple models can be integrated in the same predictive
framework as introduced by \cite{russell:17}.

\subsubsection{Path-specific Variations}
\label{sec:path-spec-vari}
A common theme of the alternatives to counterfactual fairness
\cite{kilbertus:17,ilya:18,elias:18} is the focus on
\emph{path-specific effects}, which was only briefly mentioned in our
formulation. One example for understanding path-specificity and its
relation to fairness is the previously discussed case study of gender bias in
the admissions to the University of California at Berkeley in the
1970s: gender ($A$) and admission ($Y$) were found to be associated in
the data, which lead to questions about fairness of the admission
process. One explanation found was that this was due to the choice of
department each individual was applying to ($X$). By postulating the
causal structure $A \rightarrow X \rightarrow Y$, we could claim that,
even though $A$ is a cause of $Y$, the \emph{mechanism} by which it
changes $Y$ is ``fair'' in the sense that we assume free-will in the
choice of department made by each applicant. This is of course a
judgement call that leaves unexplained why there is an interaction
between $A$ and other causes of $X$, but one that many
analysts would agree with. The problem gets more complicated if edge $A
\rightarrow Y$ is also present.

The approach by \cite{ilya:18} can tap directly from existing methods
for deriving path-specific effects as functions of $P(V)$ (see
\cite{ilya:13} for a review).  The method by \cite{kilbertus:17} and
the recent variation of counterfactual fairness by \cite{silvia:18}
consist of adding ``corrections'' to a causal model to
deactivate particular causal contributions to $Y$. This is done by
defining a particular fairness difference we want to minimise, typically the expected difference of
the outcomes $Y$ under different levels of $A$. \cite{silvia:18} suggest doing this without
parameterising a family of $\hat Y$ to be optimised. At its most basic
formulation, if we define $PSE(V)$ as the particular path-specific effect from
$A$ to $Y$ for a particular set of observed variables $V$, by taking expectations under
$Y$ we have that a fair $\hat Y$ can be simply defined as $\mathsf E[Y\ |\ V
  \backslash \{Y\}] - PSE(V)$.  Like in \cite{ilya:18}, it is however
not obvious which optimally is being achieved for the prediction of
$Y$ as it is unclear how such a transformation would translate in terms of
projecting $Y$ into a constrained space when using a particular
loss function.

\begin{figure}[!t]
\centering\includegraphics[width=2.5in]{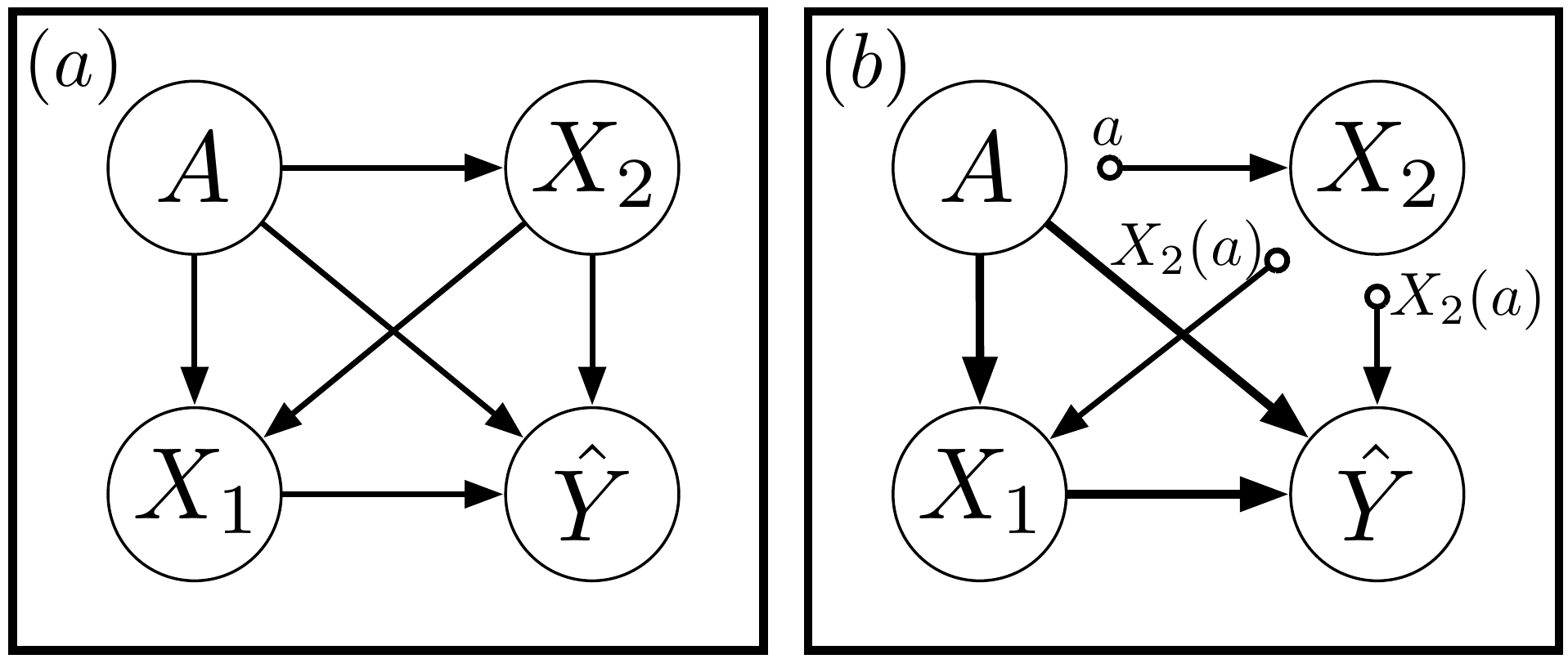}
  \caption{(a) A causal graph linking protected attribute $A$ to predictor $\hat Y$, where only a subset of edges will ``carry'' counterfactual values of $A$ in order to represent the constraints of path-specific counterfactual fairness. 
  (b) This diagram, inspired by \cite{pearl:01}, is a representation of how
  counterfactuals are propagated only through some edges. For other edges, inputs are based on the baseline value $a$ of an individual.}
\label{fig:pse}
\end{figure}

Our own suggestion for path-specific counterfactual fairness builds
directly on the original: just extract latent fair variables from observed variables that are known to be
(path-specifically) fair and build a black-box predictor around
them. For interpretation, it is easier to include $\hat Y$ in the
causal graph (removing $Y$, which plays no role as an input to $\hat Y$), adding edges from all other
vertices into $\hat Y$. Figure \ref{fig:pse}(a) shows an example with
three variables $A, X_1, X_2$ and the predictor $\hat Y$. Assume that,
similar to the Berkeley case, we forbid path 
$A \rightarrow X_1 \rightarrow \hat Y$ as an unfair contribution to $\hat Y$, while
allowing contributions via $X_2$ (that is, paths $A \rightarrow X_2
\rightarrow \hat Y$ and $A \rightarrow X_2 \rightarrow X_1 \rightarrow
\hat Y$. This generalises the Berkeley example, where $X_2$ would
correspond to department choice and $X_1$ to, say, some source of
funding that for some reason is also affected by the gender of the
applicant). Moreover, we also want to exclude the direct contribution
$A \rightarrow \hat Y$. Assuming that a unit would be set to a
factual baseline value $a$ for $A$, the ``unfair propagation'' of a
counterfactual value $a'$ of $A$ could be understood as passing it
only through $A \rightarrow X_1 \rightarrow \hat Y$ and $A \rightarrow
\hat Y$ in Figure \ref{fig:pse}(b), leaving the inputs ``through the
other edges'' at the baseline \cite{pearl:01, ilya:13}.  The relevant
counterfactual for $\hat Y$ is the \emph{nested counterfactual} $\hat
Y(a', X_1(a', X_2(a)), X_2(a))$. A direct extension of the definition
of counterfactual fairness applies to this path-specific scenario: we require
\begin{equation}
  \begin{array}{c}
  P(\hat Y(a', X_1(a', X_2(a)), X_2(a))\ |\ X_1 = x_1, X_2 = x_2, A = a) =\\
  P(\hat Y(a, X_1(a, X_2(a)), X_2(a))\ |\ X_1 = x_1, X_2 = x_2, A = a).
  \end{array}
\label{eq:pse}
\end{equation}

To enforce the constraint in (\ref{eq:pse}), the input to $\hat Y$ can be made
to depend only on the factuals which will not violate the above. For instance,
$\hat Y \equiv X_1$ will violate the above, as $\hat Y(a, X_1(a, X_2(a)), X_2(a)) =
X_1(a, X_2(a)) = x_1$ while $\hat Y(a', X_1(a, X_2(a')), X_2(a')) = X_1(a', X_2(a)) \neq x_1$
(in general).
$\hat Y \equiv X_2$ is acceptable, as  $\hat Y(a, X_1(a, X_2(a)), X_2(a))$ $=$
$\hat Y(a', X_1(a', X_2(a)), X_2(a)) = X_2(a) = x_2$.

We need a definition of $\hat Y$ that is invariant as in
(\ref{eq:pse}) to any counterfactual world we choose. Assume that, in
our example, $A \in \{a_1, \dots, a_K\}$ and the structural equation
for $X_1$ is $X_1 = f_{X_1}(A, X_2, U_{X_1})$ for some background
variable $U_{X_1}$. We can simply set
\begin{align}
\hat Y \equiv g(f_{X_1}(a_1,
X_2, U_{X_1}), \dots, f_{X_1}(a_K, X_2, U_{X_1}), X_2, U_{X_1}) \nonumber
\end{align}
for some arbitrary function $g(\cdot)$. The only descendant of $A$ in this
case is $X_2$, which we know we can use. Overall, this provides more
predictive information than what would be allowed by using $U_{X_1}$
and $U_{X_2}$ only. One important point: we know that $f_{X_1}(a, x_2,
U_{X_1}) = x_1$ when conditioning on $A = a$ and $X_2 = x_2$, which
means that $\hat Y$ is a function of $X_1$. This seems to contradict
(\ref{eq:pse}). In fact it does not, because $g(\cdot)$ is defined
\emph{without a priori knowledge of which values of $A$ will be
  factual for which units}. 
  
For instance, if $K = 2$ and $\hat Y
\equiv \alpha_{1} f_{X_1}(a_1, X_2, U_{X_1}) + \alpha_{2} f_{X_1}(a_2, X_2,
U_{X_1}) + \beta_2 X_2 + \beta_1 U_{X_1}$, then for a unit with $(A =
a_1, X_2 = x_2, U_{X_1} = u_{X_1}, X_1 = x_1 = f_{X_1}(a_1, x_2,
u_{X_1}))$ we have 
\begin{align*}
\hat Y = \alpha_{1}x_1 + \alpha_{2}f_{X_1}(a_2, x_2,
U_{X_1}) + \beta_2 x_2 + \beta_1 u_{X_1}.
\end{align*}
For a unit with $(A = a_2,
X_2 = x_2, U_{X_1} = u_{X_1}, X_1 = x_1' = f_{X_1}(a_2, x_2,
u_{X_1}))$,
\begin{align*}
\hat Y = \alpha_{1}f_{X_1}(a_1, x_2, U_{X_1}) +
\alpha_{2}x_1' + \beta_2 x_2 + \beta_1 u_{X_1}.
\end{align*}
That is, the
interaction between the factual value of $X_1$ and particular
parameters of the predictor will depend on the value of the factual
exposure $A$, not only on the functional form of $g(\cdot)$, \emph{even if
the other inputs to $\hat Y$ remain the same}. This is a
more general setup than \cite{silvia:18} which focuses on particular
effects, such as expected values, and this is a direct extension of our
original definition of counterfactual fairness.

\section{Conclusion}
Counterfactual fairness is a formal definition of fairness based on causal reasoning directly, but any mathematical formalisation of fairness can be studied within a causal framework as we have described. This approach provides tools for making the assumptions that underlie intuitive notions of fairness explicit. This is important, since notions of fairness that are discordant with the actual causal relationships in the data can lead to misleading and undesirable outcomes. Only by understanding and accurately modelling the mechanisms that propagate unfairness through society can we make informed decisions as to what should be done. 











\bibliography{bibliography,rbas}

\begin{thebibliography}{10}

\bibitem{pmlr-v81-barabas18a}
Chelsea Barabas, Madars Virza, Karthik Dinakar, Joichi Ito, and Jonathan
  Zittrain.
\newblock Interventions over predictions: Reframing the ethical debate for
  actuarial risk assessment.
\newblock In Sorelle~A. Friedler and Christo Wilson, editors, {\em Proceedings
  of the 1st Conference on Fairness, Accountability and Transparency},
  volume~81 of {\em Proceedings of Machine Learning Research}, pages 62--76,
  New York, NY, USA, 23--24 Feb 2018. PMLR.

\bibitem{elias:16}
E.~Bareinboim and J.~Pearl.
\newblock Causal inference and the data-fusion problem.
\newblock {\em Proceedings of the National Academy of Sciences},
  113:7345--7352, 2016.

\bibitem{bickel1975sex}
Peter~J Bickel, Eugene~A Hammel, and J~William O'Connell.
\newblock Sex bias in graduate admissions: Data from berkeley.
\newblock {\em Science}, 187(4175):398--404, 1975.

\bibitem{bol:89}
K.~Bollen.
\newblock {\em Structural {E}quations with {L}atent {V}ariables}.
\newblock John Wiley \& Sons, 1989.

\bibitem{bolukbasi2016man}
Tolga Bolukbasi, Kai-Wei Chang, James~Y Zou, Venkatesh Saligrama, and Adam~T
  Kalai.
\newblock Man is to computer programmer as woman is to homemaker? debiasing
  word embeddings.
\newblock In {\em Advances in Neural Information Processing Systems}, pages
  4349--4357, 2016.

\bibitem{cappelen2013just}
Alexander~W Cappelen, James Konow, Erik~{\O} S{\o}rensen, and Bertil Tungodden.
\newblock Just luck: An experimental study of risk-taking and fairness.
\newblock {\em American Economic Review}, 103(4):1398--1413, 2013.

\bibitem{silvia:18}
S.~Chiappa and T.~Gillam.
\newblock Path-specific counterfactual fairness.
\newblock {\em arXiv:1802.08139}, 2018.

\bibitem{chouldechova:17}
A.~Chouldechova.
\newblock Fair prediction with disparate impact: a study of bias in recidivism
  prediction instruments.
\newblock {\em Big Data}, 2:153--163, 2017.

\bibitem{chouldechova2017fair}
Alexandra Chouldechova.
\newblock Fair prediction with disparate impact: A study of bias in recidivism
  prediction instruments.
\newblock {\em Big data}, 5(2):153--163, 2017.

\bibitem{cohen1989currency}
Gerald~A Cohen.
\newblock On the currency of egalitarian justice.
\newblock {\em Ethics}, 99(4):906--944, 1989.

\bibitem{dawid:00}
A.~P. Dawid.
\newblock Causal inference without counterfactuals.
\newblock {\em Journal of the American Statistical Association}, 95:407--424,
  2000.

\bibitem{dwork2012fairness}
Cynthia Dwork, Moritz Hardt, Toniann Pitassi, Omer Reingold, and Richard Zemel.
\newblock Fairness through awareness.
\newblock In {\em Proceedings of the 3rd innovations in theoretical computer
  science conference}, pages 214--226. ACM, 2012.

\bibitem{dworkin2002sovereign}
Ronald Dworkin.
\newblock {\em Sovereign virtue: The theory and practice of equality}.
\newblock Harvard university press, 2002.

\bibitem{edwards2015censoring}
Harrison Edwards and Amos Storkey.
\newblock Censoring representations with an adversary.
\newblock {\em arXiv preprint arXiv:1511.05897}, 2015.

\bibitem{esteva2017dermatologist}
Andre Esteva, Brett Kuprel, Roberto~A Novoa, Justin Ko, Susan~M Swetter,
  Helen~M Blau, and Sebastian Thrun.
\newblock Dermatologist-level classification of skin cancer with deep neural
  networks.
\newblock {\em Nature}, 542(7639):115, 2017.

\bibitem{flores2016false}
Anthony~W Flores, Kristin Bechtel, and Christopher~T Lowenkamp.
\newblock False positives, false negatives, and false analyses: A rejoinder to
  machine bias: There's software used across the country to predict future
  criminals. and it's biased against blacks.
\newblock {\em Fed. Probation}, 80:38, 2016.

\bibitem{hardt2016equality}
Moritz Hardt, Eric Price, Nati Srebro, et~al.
\newblock Equality of opportunity in supervised learning.
\newblock In {\em Advances in neural information processing systems}, pages
  3315--3323, 2016.

\bibitem{he2015delving}
Kaiming He, Xiangyu Zhang, Shaoqing Ren, and Jian Sun.
\newblock Delving deep into rectifiers: Surpassing human-level performance on
  imagenet classification.
\newblock In {\em Proceedings of the IEEE international conference on computer
  vision}, pages 1026--1034, 2015.

\bibitem{kamiran2009classifying}
Faisal Kamiran and Toon Calders.
\newblock Classifying without discriminating.
\newblock In {\em Computer, Control and Communication, 2009. IC4 2009. 2nd
  International Conference on}, pages 1--6. IEEE, 2009.

\bibitem{kamishima2012fairness}
Toshihiro Kamishima, Shotaro Akaho, Hideki Asoh, and Jun Sakuma.
\newblock Fairness-aware classifier with prejudice remover regularizer.
\newblock In {\em Joint European Conference on Machine Learning and Knowledge
  Discovery in Databases}, pages 35--50. Springer, 2012.

\bibitem{kilbertus:17}
N.~Kilbertus, M.~R. Carulla, G.~Parascandolo, M.~Hardt, D.~Janzing, and
  B.~Sch\"{o}lkopf.
\newblock Avoiding discrimination through causal reasoning.
\newblock {\em Advances in Neural Information Processing Systems}, 30:656--666,
  2017.

\bibitem{kleinberg2016inherent}
Jon Kleinberg, Sendhil Mullainathan, and Manish Raghavan.
\newblock Inherent trade-offs in the fair determination of risk scores.
\newblock {\em arXiv preprint:1609.05807}, 2016.

\bibitem{kusner:17}
M.~Kusner, J.~Loftus, C.~Russell, and R.~Silva.
\newblock Counterfactual fairness.
\newblock {\em Advances in Neural Information Processing Systems},
  30:4066--4076, 2017.

\bibitem{louizos2015variational}
Christos Louizos, Kevin Swersky, Yujia Li, Max Welling, and Richard Zemel.
\newblock The variational fair autoencoder.
\newblock {\em arXiv preprint arXiv:1511.00830}, 2015.

\bibitem{lum2016predict}
Kristian Lum and William Isaac.
\newblock To predict and serve?
\newblock {\em Significance}, 13(5):14--19, 2016.

\bibitem{mollerstrom2015luck}
Johanna Mollerstrom, Bj{\o}rn-Atle Reme, and Erik~{\O} S{\o}rensen.
\newblock Luck, choice and responsibility—an experimental study of fairness
  views.
\newblock {\em Journal of Public Economics}, 131:33--40, 2015.

\bibitem{morgan:15}
S.~Morgan and C.~Winship.
\newblock {\em Counterfactuals and Causal Inference: Methods and Principles for
  Social Research}.
\newblock Cambridge University Press, 2015.

\bibitem{ilya:18}
R.~Nabi and I.~Shpitser.
\newblock Fair inference on outcomes.
\newblock {\em 32nd AAAI Conference on Artificial Intelligence}, 2018.

\bibitem{pearl:00}
J.~Pearl.
\newblock {\em Causality: {M}odels, {R}easoning and {I}nference}.
\newblock Cambridge University Press, 2000.

\bibitem{pearl:01}
J.~Pearl.
\newblock Direct and indirect effects.
\newblock {\em Proceedings of the 17th Conference on Uncertainty in Artificial
  Intelligence}, pages 411--420, 2001.

\bibitem{richard1988equality}
J~Arneson Richard.
\newblock Equality and equal opportunity for welfare.
\newblock In {\em Theories of Justice}, pages 75--91. Routledge, 1988.

\bibitem{richardson:13}
T.S. Richardson and J.~Robins.
\newblock Single world intervention graphs ({SWIG}s): A unification of the
  counterfactual and graphical approaches to causality.
\newblock {\em Working Paper Number 128, Center for Statistics and the Social
  Sciences, University of Washington}, 2013.

\bibitem{robins:86}
J.~Robins.
\newblock A new approach to causal inference in mortality studies with a
  sustained exposure period-application to control of the healthy worker
  survivor effect.
\newblock {\em Mathematical Modelling}, 7:1395--1512, 1986.

\bibitem{rubin:74}
D.~Rubin.
\newblock Estimating causal effects of treatments in randomized and
  nonrandomized studies.
\newblock {\em Journal of Educational Psychology}, 66:688--701, 1974.

\bibitem{russell:17}
C.~Russell, M.~Kusner, J.~Loftus, and R.~Silva.
\newblock When worlds collide: integrating different counterfactual assumptons
  in fairness.
\newblock {\em Advances in Neural Information Processing Systems},
  30:6417--6426, 2017.

\bibitem{ilya:13}
I.~Shpitser.
\newblock Counterfactual graphical models for longitudinal mediation analysis
  with unobserved confounding.
\newblock {\em Cognitive Science}, 32:1011--1035, 2013.

\bibitem{silver2016mastering}
David Silver, Aja Huang, Chris~J Maddison, Arthur Guez, Laurent Sifre, George
  Van Den~Driessche, Julian Schrittwieser, Ioannis Antonoglou, Veda
  Panneershelvam, Marc Lanctot, et~al.
\newblock Mastering the game of go with deep neural networks and tree search.
\newblock {\em nature}, 529(7587):484--489, 2016.

\bibitem{simpson1951interpretation}
Edward~H Simpson.
\newblock The interpretation of interaction in contingency tables.
\newblock {\em Journal of the Royal Statistical Society. Series B
  (Methodological)}, pages 238--241, 1951.

\bibitem{sgs:93}
P.~Spirtes, C.~Glymour, and R.~Scheines.
\newblock {\em Causation, {P}rediction and {S}earch}.
\newblock Lecture Notes in Statistics 81. Springer, 1993.

\bibitem{spirtes:95b}
P.~Spirtes, C.~Meek, and T.~Richardson.
\newblock Causal inference in the presence of latent variables and selection
  bias.
\newblock {\em Proceedings of the 11th International Conference on Uncertainty
  in Artificial Intelligence (UAI 1995)}, pages 499--506, 1995.

\bibitem{sweeney2013discrimination}
Latanya Sweeney.
\newblock Discrimination in online ad delivery.
\newblock {\em Queue}, 11(3):10, 2013.

\bibitem{zafar2017fairness}
Muhammad~Bilal Zafar, Isabel Valera, Manuel Gomez~Rodriguez, and Krishna~P
  Gummadi.
\newblock Fairness constraints: Mechanisms for fair classification.
\newblock {\em arXiv preprint arXiv:1507.05259}, 2017.

\bibitem{zemel2013learning}
Rich Zemel, Yu~Wu, Kevin Swersky, Toni Pitassi, and Cynthia Dwork.
\newblock Learning fair representations.
\newblock In {\em International Conference on Machine Learning}, pages
  325--333, 2013.

\bibitem{elias:18}
J.~Zhang and E.~Bareinboim.
\newblock Fairness in decision-making: the causal explanation formula.
\newblock {\em 32nd AAAI Conference on Artificial Intelligence}, 2018.

\end{thebibliography}
\bibliographystyle{plain}

\end{document}